\documentclass[twoside,11pt]{article}

\usepackage{blindtext}

\usepackage{jmlr2e}

\usepackage{amssymb}
\usepackage{amsmath}

\usepackage{booktabs}
\usepackage{geometry}
\usepackage{adjustbox}
\usepackage{multirow}
\usepackage{pifont}
\usepackage{natbib}
\usepackage{tabularx}

\newcommand{\cmark}{\checkmark}
\newcommand{\xmark}{\ding{55}}

\usepackage{float}      
\usepackage{caption}    
\usepackage{subcaption} 

\usepackage{hyperref}
\usepackage{xcolor}

\usepackage{graphicx}
\usepackage{array}
\usepackage{booktabs}
\usepackage{ragged2e}
\usepackage{caption}

\usepackage{listings}
\definecolor{codegreen}{rgb}{0,0.6,0}
\definecolor{codegray}{rgb}{0.5,0.5,0.5}
\definecolor{codepurple}{rgb}{0.58,0,0.82}
\definecolor{backcolour}{rgb}{0.97,0.97,0.97}

\lstdefinestyle{mystyle}{
    backgroundcolor=\color{backcolour},   
    commentstyle=\color{codegreen},
    keywordstyle=\color{blue},
    numberstyle=\tiny\color{codegray},
    stringstyle=\color{red},
    basicstyle=\ttfamily\small,
    breaklines=true,
    captionpos=b,
    keepspaces=true,
    numbers=left,
    numbersep=5pt,
    showspaces=false,
    showstringspaces=false,
    showtabs=false,
    tabsize=2
}

\newcommand{\ben}[1]{\textcolor{blue}{\textbf{Ben:} #1}}

\usepackage[most]{tcolorbox} 
\usepackage{xcolor}          

\usepackage{lastpage}
\jmlrheading{X}{2025}{1-\pageref{LastPage}}{9/25; Revised X/XX}{X/XX}{XX-0000}{Shamir et al.}

\ShortHeadings{Online Dynamic GR in Gym Envs}{Shamir et al.}
\firstpageno{1}

\begin{document}



\title{Online Dynamic Goal Recognition in Gym Environments} 

\author{
\name \hspace{-3mm} Matan Shamir \email matan.shamir@live.biu.ac.il \\ \name Osher Elhadad \email osher.elhadad@live.biu.ac.il \\ \name Ben Nageris \email benabraham.nageris@live.biu.ac.il
 \\
\addr Computer Science, Bar-Ilan University\\
Ramat Gan, Israel
\AND
\name Reuth Mirsky \email reuth.mirsky@tufts.edu \\
\addr Computer Science, Tufts University\\
Medford, MA, USA
}
\editor{My editor}

\maketitle

\begin{abstract}
Goal Recognition (GR) is the task of inferring an agent's intended goal from partial observations of its behavior, typically in an online and one-shot setting. 
Despite recent advances in model-free GR, particularly in applications such as human-robot interaction, surveillance, and assistive systems, the field remains fragmented due to inconsistencies in benchmarks, domains, and evaluation protocols.

To address this, we introduce \texttt{gr-libs}\footnote{\url{https://github.com/MatanShamir1/gr_libs}}  and \texttt{gr-envs}\footnote{\url{https://github.com/MatanShamir1/gr_envs}}, two complementary open-source frameworks that support the development, evaluation, and comparison of GR algorithms in Gym-compatible environments. \texttt{gr-libs} includes modular implementations of MDP-based GR baselines, diagnostic tools, and evaluation utilities. \texttt{gr-envs} provides a curated suite of environments adapted for dynamic and goal-directed behavior, along with wrappers that ensure compatibility with standard reinforcement learning toolkits. Together, these libraries offer a standardized, extensible, and reproducible platform for advancing GR research. Both packages are open-source and available on GitHub and PyPI.
\end{abstract}





\section{Introduction}

Goal Recognition (GR) is the task of inferring an actor's intended goal from a sequence of observed behaviors within a given domain.  It plays a central role in domains such as human-robot interaction~\citep{levine2014concurrent}, surveillance~\citep{ang2017game, mirsky2019new}, tutoring~\citep{amir2011plan,gupta2022enhancing,wilson2022help}, assistive technologies~\citep{huntemann2007bayesian,jiang2021goal}, and gaming~\citep{synnaeve2011bayesian,min2017multimodal}. Despite its breadth, GR research remains fragmented due to divergent assumptions about input formats, actor models, and evaluation criteria, hindering reproducibility and fair comparison. Most of the problem formulations used for GR comply with the following definition \citep{sukthankar2014plan,meneguzzi2021survey,mirsky2021symbolic,amado2024survey}:

\vspace{-2mm}
\begin{definition}
A \textbf{Goal Recognition problem} is a tuple $\langle T, G, O\rangle$, where $T$ is the domain theory, $G$ is the potential goal space, and $O$ is a sequence of observations. The output is a goal $g \in G$ that best explains $O$.
\end{definition}
\vspace{-2mm}

This definition leaves the concrete description of $T$ and how to explain $O$ through $G$, as they vary between GR approaches.
Furthermore, this representation assumes static goal sets, which pose a challenge for using GR in realistic environments, where goals might change over time. Online Dynamic Goal Recognition (ODGR) \citep{shamir2024RLC} extends GR by introducing a phased structure in which new goal sets and observation sequences arrive over time: first the algorithm is introduced to the underlying environment representation ($T$), then a potential set of goals the actor might be pursuing ($G$) and then an obervation sequence ($O$), denoting a single GR problem to solve. Formally, 

\vspace{-2mm}
\begin{definition}
An \textbf{Online Dynamic Goal Recognition (ODGR)} problem is a tuple 
$\langle T, \langle G^i, \{O\}^i \rangle_{i\in 1..n}\rangle$, 
where $T = \langle S, A \rangle$ is a domain theory, 
$G^i$ is a set of goals, and $\{O\}^i$ is a set of observation sequences. Each observation sequence $O^i_j = \langle \langle s_{j,1}^i, a_{j,1}^i \rangle, \langle s_{j,2}^i, a_{j,2}^i \rangle, \ldots \rangle$ arrives after $G^i$, and the recognizer must return the most likely goal $g_j^i \in G^i$. The final output is a collection $G^* = \{\{g_1^1, g_2^1, ...\}, \{g_1^2, g_2^2, ...\}, ..., \{g_1^n, g_2^n, ...\}\}$.
\end{definition}
\vspace{-2mm}

\begin{figure}[ht]
  \centering
  \makebox[\textwidth][c]{%
    \includegraphics[width=\textwidth]{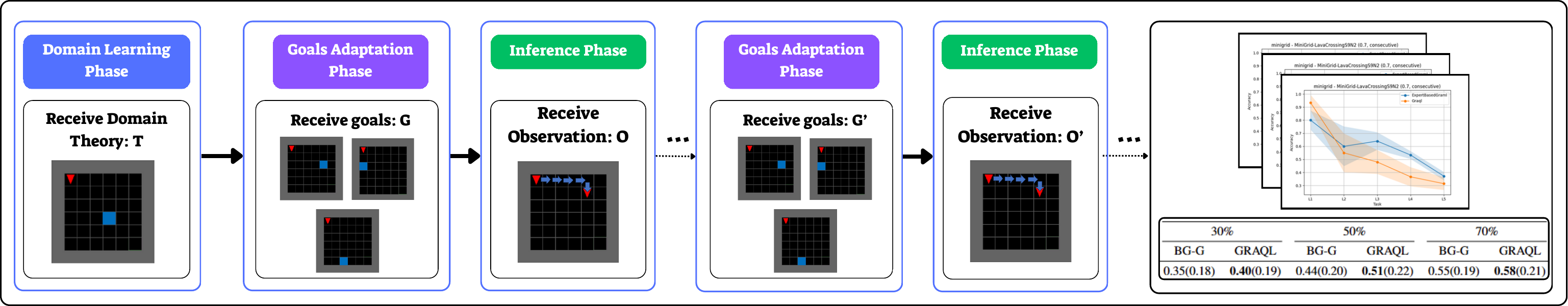}
  }
  \caption{Steps of execution of an ODGR problem by the problem execution engine, including outputs and statistics over the execution.}
  \label{fig:GRAML}
\end{figure}

The ODGR formulation explicitly separates the ODGR process into three main phases: (1) domain learning, (2) goal adaptation, and (3) inference intervals. Figure \ref{fig:GRAML} illustrates the hierarchical structure of ODGR phases. After an initial full cycle of domain learning, goal adaptation, and inference, the process can re-enter at different stages: a new cycle may start with (2–3), (3) alone, or a full (1–2–3), depending on the incoming data. This re-entrant phased structure allows the system to adapt to changing goals and perform inference under evolving conditions.

\begin{tcolorbox}[colback=gray!10,colframe=black!70]
To consolidate work on ODGR and facilitate rapid research progress, we introduce two open-source frameworks: \texttt{gr-libs}, a modular library of GR algorithms and benchmarking tools, and \texttt{gr-envs}, a suite of Gym-compatible environments specifically designed and modified for ODGR tasks.
\end{tcolorbox}

Together, these packages support GR and ODGR reproducible experimentation, standardized environments for benchmarking, goal-conditioned behavior generation for policy learning and observation sequences, and systematic evaluation across diverse settings.

\paragraph{Related Work}
Goal Recognition (GR) has long been studied under various names and as part of other problems, including plan recognition~\citep{kautz1986generalized}, intention recognition~\citep{cohen1990intention}, and inverse planning~\citep{ng2000algorithms, baker2009action}.
Symbolic approaches such as Plan Recognition as Planning~\citep{RamirezGeffner09,ramirez2010probabilistic} and its extensions~\citep{masters2017cost, vered2017heuristic, sohrabi2016plan, pereira2020landmark, meneguzzi2021survey} rely on handcrafted domain models, limiting scalability in dynamic settings. Learning-based methods like GRNet~\citep{chiari2022goal,chiari2024fast} inherit strong representational assumptions from symbolic planning. This benchmark focuses on model-free techniques \citep{min2014deep, polyvyanyy2020goal,shvo2021interpretable}, especially RL-based such as GRAQL~\citep{amado2022goal} and its successors~\citep{fang2023real,nageris2024goal,wen2025goal}, which require extensive training and are sensitive to domain shifts, yet their evaluation pipelines are often bespoke and lack standardization.

\vspace{-3mm}
\section{gr-libs}
\vspace{-2mm}
\texttt{gr-libs} provides a structured implementation of ODGR algorithms within an MDP-based setting.
To enable systematic evaluation of GR algorithms under controlled yet diverse conditions, we provide a structured benchmark of ODGR problems within the \texttt{gr$\_$libs} package. These problems are defined in a dedicated configuration file, \texttt{consts.py}, and follow a standardized format that reflects the structure of an ODGR problem. The benchmark is designed to give all algorithms a fair opportunity to demonstrate their strengths and weaknesses, spanning variations in properties such as degree of observability, input sequence optimality, and whether the sequence is consecutive or non-consecutive.

The benchmark includes a growing suite of baseline GR algorithms. All implemented algorithms are fully compatible with the \texttt{gr-envs} and \texttt{gr-libs} infrastructure, which standardizes environments via the Gymnasium API and agent training via SB3. This ensures reproducibility and comparability across diverse domains.
Every implemented algorithm can be executed only on environments whose properties qualify for the algorithm's requirements. For example, since GRAQL is an algorithm that can only work in discrete environments, it cannot be executed in PointMaze, which is a continuous domain. 
Table \ref{tab:algos} specifies the properties required by every algorithm in \texttt{gr-libs}.

\begin{table}[ht]
\centering
\scriptsize
\caption{Comparison of supported algorithms by space compatibility, goal generalization capability, and GC environment requirements (specified in Appendix \ref{app:convert})}
\label{tab:algos}
\begin{tabular}{l l l l}
\toprule
\textbf{Algorithm} & \textbf{State/Action Space} & \textbf{Practical Adaptation to New Goals} & \textbf{Requires GC-Adaptable Env} \\
\midrule
GRAQL                & Discrete              & \xmark~ & \xmark \\
DRACO                & Discrete/Continuous   & \xmark~ & \xmark \\
GC-DRACO             & Discrete/Continuous   & \cmark~ & \cmark \\
GC-AURA              & Discrete/Continuous   & \cmark~ & \cmark \\
BG-GRAML             & Discrete/Continuous   & \cmark~ & \xmark \\
GC-GRAML             & Discrete/Continuous   & \cmark~ & \cmark \\
\bottomrule
\end{tabular}
\end{table}

\vspace{-3mm}
\section{gr-envs}
\vspace{-2mm}
The benchmark suite includes a curated collection of environments covering both discrete and continuous domains.
\texttt{gr-envs} provides structural adaptations of Gymnasium environments to meet the requirements of GR algorithms. Through gym-wrappers, specialized initialization methods, and custom classes, the package ensures that the algorithms in \texttt{gr-libs} can run smoothly and receive all necessary information in a consistent representation.
In addition, \texttt{gr-envs} offers tooling for debugging and visualization. For example, custom wrappers overlay color-coded traces in environments such as Minigrid to indicate past agent actions. We also include scripts for rendering episodes to image or video formats. When run with a debug flag, recognizers can export full visual traces of observed trajectories to a specified output directory, simplifying the qualitative analysis and error inspection process.

\section{Conclusions and Future Work} 
\label{sec:conclusions}
This paper introduces \texttt{gr-libs} and \texttt{gr-envs}, two open-source packages that standardize algorithms and environments for Online Dynamic Goal Recognition.
Additional resources can be found in the appendices, and include more elaborate information about the ODGR problem and its phases (Appendix \ref{app:phases}), an overview on how to run an ODGR problem (Appendix \ref{app:exec_overview}) and specific walkthroughts for manual execution (Appendix \ref{app:manual-odgr-example}) and using a script (Appendix \ref{app:engine-odgr-example}), usability guidelines (Appendix \ref{app:usability}) and a list of supported algorithms (Appendix \ref{app:algs}) and environments (Appendix \ref{app:envs}). Further, for researchers who are interested in extending this work to new environments, Appendix \ref{app:convert} provides guidelines on converting a Gymnasium-compatible environment into an ODGR environment. Finally, for benchmarking and comparison of novel work to the state-of-the-art, Appendix \ref{app:eval} provides some baselines and standard tests.
By providing reproducible benchmarks, modular implementations, and Gymnasium-compatible environments, our work enables fair comparison across methods and accelerates progress in model-free GR. We hope this foundation will foster broader adoption, easier benchmarking, and new research on dynamic goal recognition in realistic settings.

\bibliography{cas-refs}


\section*{Licensing and Responsibility}
The source code for \texttt{gr-libs} and \texttt{gr-envs} is released under the MIT license.
The authors bear all responsibility in case of violation of rights in the proposed environments and included benchmark.

\appendix

\section{ODGR Phases}
\label{app:phases}

\begin{figure}[b]
  \centering
  \includegraphics[width=0.7\textwidth]{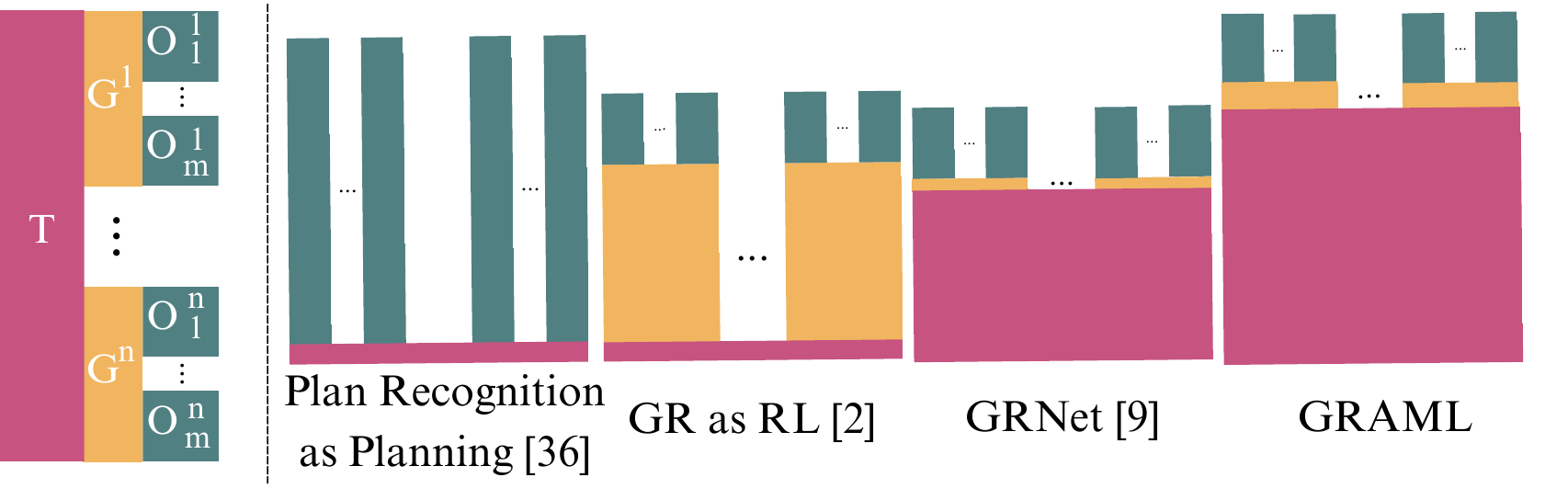}
  \caption{Depiction of ODGR intervals and the inputs at each (left) and a rough approximation of the time spent on each phase by representative GR frameworks (right).}
  \label{fig:time_spent}
\end{figure}

Figure~\ref{fig:time_spent} illustrates the structure and timing of ODGR phases.

Each ODGR problem in the benchmark includes a domain learning phase, and a single goal adaptation phase followed by multiple inference phases. In the adaptation phase, a new set of goals, distinct from those used during domain learning, is introduced.

Following adaptation, the system enters the inference phase. For each goal in the active set, a series of test traces is presented to the recognizer. These traces are systematically varied across three axes:
\begin{itemize}
    \item \textbf{Observability level}: Observation sequences are truncated to 30\%, 50\%, 70\%, or 100\% of the full trajectory.
    \item \textbf{Trace type}: Sequences are either \emph{consecutive} (prefixes of trajectories) or \emph{non-consecutive} (randomly sampled subsets).
    \item \textbf{Trace optimality}: A stochastic policy generates sub-optimal observation sequences by adding gaussian noise to the states and actions values.
\end{itemize}

All ODGR problems are defined in the configuration file \texttt{consts.py}. Each ODGR problem is associated with a specific gymnasium environment, and includes a base goal set $\bar{G}$ and a set of 3 to 9 active goals $G^0$ used in a single goal adaptation phase. For each goal $g \in G^0$, a dedicated recognition problem is created, resulting in approximately 200 GR problems per ODGR instance on average. These problems systematically vary along key axes: goal identity, observability level, and trace type (5 goals × 4 observability levels × 2 trace types), producing 40 inference phases per ODGR instance. On average, five ODGR problems are defined per environment, yielding hundreds of structured recognition problems. This setup enables fine-grained performance analysis across diverse conditions and promotes reproducibility and fair comparison by explicitly controlling goal shifts, agent behavior, and observation constraints within a robust and extensible benchmark.

\section{ODGR Problem Execution -- Overview}
\label{app:exec_overview}

Experiments can be defined and executed using the tools provided by the \texttt{gr-libs} package. The framework supports two primary modes of setup: manual scripting and declarative configuration files.

\paragraph{Manual Setup}
Users can manually construct experiments by programmatically instantiating environments, defining Goal Recognition algorithms, and executing recognition routines using the core API provided by \texttt{gr-libs}. This approach enables full flexibility in modifying environment wrappers, controlling inference behavior, and customizing goal dynamics for tailored experiments or debugging. Appendix \ref{app:manual-odgr-example} provides a concrete example of manually executing an ODGR problem using the MiniGrid domain.
Notably, the benchamrk problems in \texttt{consts.py} can be used manually.

\paragraph{Configuration Files}
To simplify large-scale experiments, the framework has an engine that can execute an ODGR problem from the benchmark and produce a summary report before it exists. These specifics that should be specified are the environment, agent training parameters, goal recognition settings, and evaluation metrics.

When executed, the engine of the framework processes each GR problem and stores its output. Each output record is outputted in a textual and pickled format, including the inferred goal for each observation sequence, recognition probabilities (if applicable), and evaluation metrics, such as accuracy, inference time, and rank.

Results are stored in structured directories, optionally with image or video visualizations of the agent’s trajectory overlaid on the environment grid. These outputs facilitate systematic evaluation, comparison, and debugging of GR algorithms.
Appendix ~\ref{app:engine-odgr-example} shows usage example.

\paragraph{Parallel Execution}
The framework supports parallel execution of GR problems across multiple CPU cores. When enabled, each process evaluates a distinct goal recognition query independently, allowing benchmarks to scale efficiently across large problem sets. This is particularly useful for evaluating algorithm performance over different observability levels, goals, and environments.

\paragraph{Visualization tools}
The script \texttt{generate\_experiment\_results.py} processes the outputs from \texttt{odgr\_executor.py}, which executes a specific algorithm on a given ODGR task. It then plots the average accuracy across selected tasks for multiple algorithms within a chosen environment.

\begin{figure}[H]
  \centering
  \begin{subfigure}[b]{0.48\textwidth}
    \includegraphics[width=\textwidth]{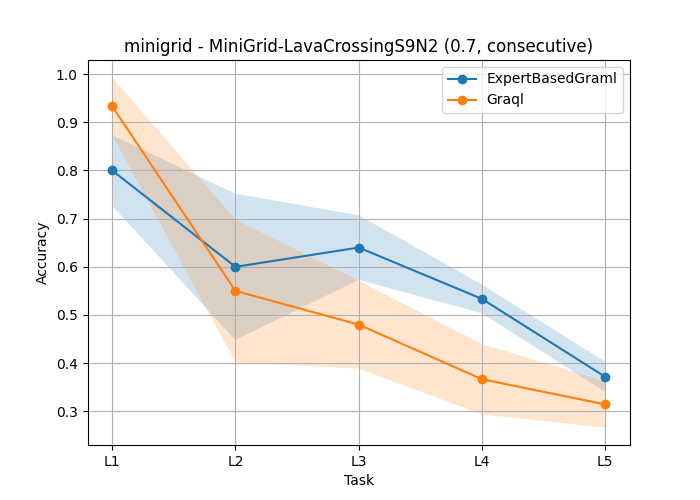}
    \caption{ExpertBasedGraml (a BG‑GRAML variant) vs.\ Graql in MiniGrid‑LavaCrossingS9N2. Observations truncated to 70\% (consecutive).}
    \label{fig:graql_graml_comp}
  \end{subfigure}
  \hfill
  \begin{subfigure}[b]{0.48\textwidth}
    \includegraphics[width=\textwidth]{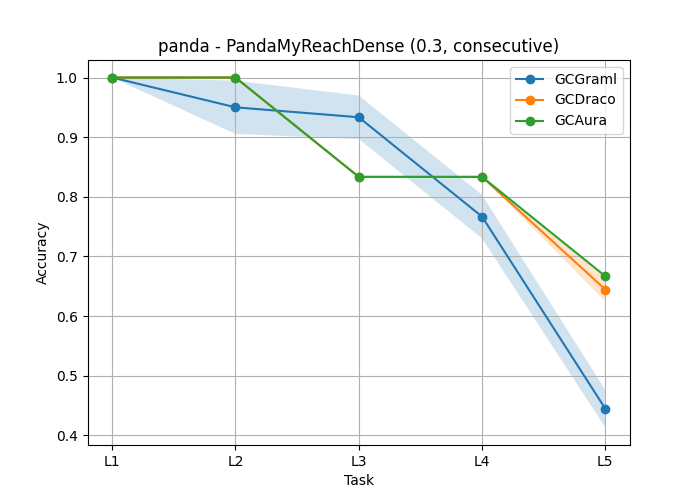}
    \caption{GCAura, GCDraco and GCGraml in PandaMyReachDense. Observations truncated to 30\% (consecutive).}
    \label{fig:aura_draco_graml_comp}
  \end{subfigure}
  \caption{Comparison of GR algorithms across two representative environments.}
  \label{fig:exp_comparison}
\end{figure}

Figures~\ref{fig:exp_comparison}a and \ref{fig:exp_comparison}b show representative outputs.

\section{Manual Execution Walkthrough}
\label{app:manual-odgr-example}

This appendix provides a minimal working example of how to manually configure and execute an Online and Dynamic Goal Recognition (ODGR) problem using the \texttt{gr-libs} interface. The example is based on the MiniGrid domain.

\vspace{1em}
\noindent
In this example:
\begin{itemize}
  \item The environment used is \texttt{MiniGrid-SimpleCrossingS13N4}.
  \item The set of dynamic goals used during the goal adaptation phase is \texttt{\{(11,1), (11,11), (1,11)\}}.
  \item The base goal set $\bar{G}$ is implicitly defined as all goals used in inference.
  \item A Tabular Q-learning agent is trained towards goal \texttt{(11,1)}.
  \item A partial observation sequence is generated from the agent's behavior using random sampling.
  \item Inference is performed using the Graql recognizer to estimate the agent's intended goal.
\end{itemize}

\vspace{1em}
\noindent
\textbf{Code Listing:}

\begin{lstlisting}[language=Python, caption={Manual ODGR problem execution using gr-libs in MiniGrid.}, label={lst:manual-odgr}]
from gr_libs import Graql
from gr_libs.environment._utils.utils import domain_to_env_property
from gr_libs.environment.environment import MINIGRID, QLEARNING
from gr_libs.metrics.metrics import stochastic_amplified_selection
from gr_libs.ml.tabular.tabular_q_learner import TabularQLearner
from gr_libs.ml.utils.format import random_subset_with_order

def run_graql_minigrid_tutorial():
    recognizer = Graql(domain_name="minigrid", env_name="MiniGrid-SimpleCrossingS13N4")

    # No training required for Graql recognizer
    recognizer.goals_adaptation_phase(
        dynamic_goals=[(11, 1), (11, 11), (1, 11)],
        dynamic_train_configs=[(QLEARNING, 100000) for _ in range(3)],
    )

    property_type = domain_to_env_property(MINIGRID)
    env_property = property_type("MiniGrid-SimpleCrossingS13N4")

    actor = TabularQLearner(
        domain_name="minigrid",
        problem_name="MiniGrid-SimpleCrossingS13N4-DynamicGoal-11x1-v0",
        env_prop=env_property,
        algorithm=QLEARNING,
        num_timesteps=100000,
    )
    actor.learn()

    full_sequence = actor.generate_observation(
        action_selection_method=stochastic_amplified_selection,
        random_optimalism=True,
    )

    partial_sequence = random_subset_with_order(
        full_sequence, int(0.5 * len(full_sequence)), is_consecutive=False
    )

    closest_goal = recognizer.inference_phase(partial_sequence, (11, 1), 0.5)
    print(f"closest_goal returned by Graql: {closest_goal}")
    print("actual goal actor aimed towards: (11, 1)")
\end{lstlisting}

\vspace{1em}
\noindent
This example demonstrates how to use \texttt{gr-libs} directly to simulate a complete ODGR episode. The recognizer is initialized and adapted to a dynamic goal set, and inference is tested with partial and noisy observations. This process supports debugging, custom evaluation, and research extensions beyond the predefined benchmark protocol.

\section{Execution Using \texttt{odgr\_executor.py} Walkthrough}
\label{app:engine-odgr-example}
This appendix demonstrates how to run a predefined Online and Dynamic Goal Recognition (ODGR) problem using the script \texttt{odgr\_executor.py}. The script executes a benchmark configuration from the \texttt{PROBLEMS} dictionary defined in \texttt{gr\_libs.problems.consts}, running all relevant phases: domain learning (if needed), goals adaptation, and inference.

\vspace{1em}
\noindent
To run the script from the command line, use:

\begin{lstlisting}[language=bash, caption={Example command to run a predefined ODGR benchmark problem.}, label={lst:odgr_executor_command}]
python odgr_executor.py \
  --domain parking \
  --env_name Parking-S-14-PC--v0 \
  --recognizer GCGraml \
  --task L1 \
  --experiment_num 0 \
  --collect_stats
\end{lstlisting}

\noindent
This command will execute all GR problem instances defined under task \texttt{L1} for the specified environment and recognizer. It will apply goal adaptation and inference across different conditions including:
\begin{itemize}
    \item Observation sequence type: consecutive vs. non-consecutive
    \item Observation completeness: varying percentages (e.g., 30\%, 50\%, ..., 100\%)
    \item Noise in agent behavior: \ben{is it relevant? Maybe we can keep until the ":" }introduced during sequence generation
\end{itemize}

\vspace{0.5em}
\noindent
The script produces a dictionary of results that includes:
\begin{itemize}
    \item Accuracy per condition (percentage $\times$ sequence type)
    \item Average inference time
    \item Total correct recognitions and overall accuracy
    \item Goal adaptation and domain learning durations
\end{itemize}

\vspace{1em}
\noindent
The output is saved automatically in:
\begin{lstlisting}[language=bash]
outputs/<recognizer>/<domain>/<env_name>/<task>/experiment_results/
\end{lstlisting}

\vspace{0.5em}
\noindent
For example, the output for the command above is saved in:

\begin{lstlisting}[language=bash]
outputs/GCGraml/parking/Parking-S-14-PC--v0/L1/experiment_results/res_0.txt
\end{lstlisting}

\vspace{1em}
\noindent
An excerpt from the resulting output:
\begin{lstlisting}[language=Python]
{
 '0.3': {'consecutive': {'accuracy': 0.33}, 'non_consecutive': {'accuracy': 1.0}},
 '0.5': {'consecutive': {'accuracy': 0.67}, 'non_consecutive': {'accuracy': 1.0}},
 ...
 '1': {'consecutive': {'accuracy': 1.0}, 'non_consecutive': {'accuracy': 1.0}},
 'total': {
   'total_correct': 27,
   'total_tasks': 30,
   'total_accuracy': 0.9,
   'total_average_inference_time': 0.022,
   'goals_adaptation_time': 12.17,
   'domain_learning_time': 15.30
 }
}
\end{lstlisting}

\noindent
This tool provides a reproducible and automated method to run full benchmark evaluations on predefined ODGR setups, making it suitable for both large-scale experiments and fine-grained performance analysis.

\section{Usability Guidelines}
\label{app:usability}
To ease usage of the package, \texttt{gr-libs} can aid datasets of pre-trained agents, pre-trained algorithm weights and existing docker images.

\subsection{Pre-trained agents}
The framework includes a pre-generated dataset of trained agents corresponding to the problems defined in the benchmark configuration file. For each supported environment and goal set, goal-conditioned policies have been pre-trained and stored, enabling immediate use during evaluation without requiring repeated training. This repository of agents significantly accelerates experimentation by allowing researchers to focus solely on developing recognition methodologies, and ensures consistent behavior generation across runs, supporting reproducibility and faster development cycles.
These sets of pre-trained agents and networks can be achieved in one of 2 ways:
1. from a Docker image in a GitHub repository. 
2. From a public google drive link. 

The pre-trained agents, used both to generate observation sequences and to support all algorithms in \texttt{gr-libs} during specific phases, were trained as follows: each single-goal agent for $300000$ timesteps, and each goal-conditioned agent for $1$ million timesteps.

\subsection{\texttt{gr$\_$cache} collection}
Certain baselines require considerable training during the domain learning phase, irrespective of agent training. For example, Graml variants construct a dataset of sequences generated by RL agents and train an LSTM on this data. To facilitate direct comparison with Graml without the need for retraining, we provide a cache directory, \texttt{gr$\_$cache}, which stores these precomputed sequences.

\section{\texttt{gr-libs} Supported Algorithms}
\label{app:algs}
The benchmark currently supports the following algorithms:

\begin{itemize}
    \item \textbf{GRAQL}~\cite{amado2022goal} – A tabular Q-learning based GR method that learns a Q-function per goal and uses similarity measures (e.g., utility, KL-divergence) to infer the most likely goal. Originally designed for environments with static goals spaces, requiring extensive agent training in case goals change. Compatible only with discrete state and action spaces.
    \item \textbf{DRACO}~\cite{nageris2024goal} – Deep Reinforcement Learning framework for GR in both discrete and continuous spaces, operating directly on raw observations without requiring symbolic input nor discretization. DRACO consists of two phases: \textit{offline learning}, where a goal-directed neural policy is trained for each candidate using an actor-critic method, and \textit{online inference}, where it estimates the likelihood that each trained policy could have produced the observed behavior. The goal candidate with the highest likelihood is selected as the agent’s intended goal. DRACO also introduces two fast, statistically grounded metrics to assess the alignment between a policy and an observation sequence, supporting partial, full, and noisy inputs and continuous environments. While compatible with both discrete and continuous action and state spaces, DRACO also (like GRAQL) assumes a fixed goal set and is not designed for dynamically changing goals.
    \item \textbf{GC-DRACO} – An extension to DRACO that uses a single goal-conditioned neural policy to cover the entire goal space, significantly reducing computational overhead and memory usage. This design removes the need to train one policy per goal, as in DRACO and GRAQL, and enables practical scalability to environments with a high number of possible goals or frequently changing goal sets. GC-DRACO generalizes across goals by adding the goal to the state representation, allowing the model to adapt to unseen or newly added goals at inference time without retraining. It also decouples goal specification from training: the goal candidates list is only required in the \textit{inference phase}, unlike DRACO and GRAQL, which require it in the \textit{offline learning} phase. This flexibility makes GC-DRACO suitable for dynamic, real-world applications where the goal space is large, continuous, or not known in advance.

    \item \textbf{GC-AURA} – An implementation of the Adaptive Universal Recognition Algorithm (AURA)~\cite{elhadad2025generaldynamicgoalrecognition} based on GCRL. GC-AURA trains a single goal-conditioned policy that generalizes across the goal space (or subspace), enabling recognition without retraining from scratch for each candidate. Unlike DRACO and GRAQL, which require a fixed goal set during offline learning, GC-AURA decouples goal specification from training, requiring only the candidate set at inference time. This allows it to scale to dynamic environments with continuous or previously unseen goals. In \texttt{gr-libs}, GC-AURA further supports recognition over \textit{goal subspaces}, avoiding the infeasibility of training over the entire (potentially huge) goal space. To address underrepresented or difficult goals, GC-AURA allows flexible adaptation strategies: \textit{zero-shot transfer}, \textit{goal recall} (retrieving stored policies), and \textit{few-shot fine-tuning} of the GCRL policy into a specific goal-directed policy. This fine-tuning capability relaxes the unrealistic assumption that a single GCRL policy is perfect for all goals, while still retaining scalability and efficiency. In addition, it is compatible with both discrete and continuous state and action spaces. Experimental results show that GC-AURA achieves robust recognition under noise and partial observability, and provides faster inference than goal-specific training approaches, making it suitable for real-world dynamic GR settings.
    \item \textbf{GRAML} \cite{shamir2025gramldynamicgoalrecognition} – A self-supervised, metric learning–based method that performs recognition in an embedding space rather than in the state space. Given an MDP, GRAML selects a set of prototype goals, trains goal-directed policies towards them, and generates samples for metric learning, comprised of 2 sequences of observations labeled with 1 if they lead to the same goal, and 0 otherwise. To adapt to possible emerging goals, GRAML collects goal-directed sequences per possible goal, and in inference time, it embeds the input sequence along with the goal-directed sequences per goal GRAML is designed for realistic environments and can operate effectively in highly dynamic, continuous settings. Because GRAML controls the composition of the dataset used to train its metric model, it can recognize suboptimal and noisy behavior by deliberately injecting such behavior into the dataset.
    \begin{itemize}
      \item \textbf{BG-GRAML} – A GRAML variant which uses goal-directed policies towards a selected set of prototype goals. In the goal adaptation phase, BG-GRAML has to receive goal-directed sequences per new goal, or it employs an MDP planner such as MCTS to generate the required example traces.
      \item \textbf{GC-GRAML} – A GRAML variant which uses a goal-conditioned policy during domain learning. It then eliminates the need for additional traces in the goal adaptation phase, because it can generate them autonomously, showing better empirical accuracy and stability.
  \end{itemize}
\end{itemize}

\section{Supported Environments in \texttt{gr-envs}}
\label{app:envs}
We elaborate on the specifics of every supported environment in \texttt{gr-envs}. A summary of the environments properties along with representing images is provided in Table \ref{tab:domains}.

\textbf{Minigrid}
consists of two distinct custom environments: Minigrid LavaCrossing  and Minigrid SimpleCrossing environments. In both scenarios, the agent (represented by the red arrow) must reach the green goal square within a grid world, starting each episode at the grid square (1,1). The first environment is a $13 \times 13$ grid featuring walls along its borders and at various points in the middle. 
The presence of goals behind walls may lead to diverse optimal plans from the starting position of the agent. The second environment is a 9x9 grid with two lava crossings in the middle. If the agent touches the lava, it incurs a negative reward and a termination of the episode. In both environments, the average number of steps taken per episode is around ten steps.

\textbf{PointMaze}
We develop two distinct custom environments: PointMaze-Obstacle and PointMaze-FourRooms. In both scenarios, the agent, represented by a 2-DoF ball force-actuated in the Cartesian directions (x and y), must reach a goal within a closed maze, starting each episode from the top-left corner. The PointMaze-FourRooms environment is a classic setup featuring four rooms, with walls along its borders and various locations throughout the grid. 
Goals in the bottom-right room may lead to diverse optimal plans originating from the starting position of the agent. In the PointMaze-Obstacle environment, a square object is placed in the center of the grid, obstructing the path of the agent and ensuring that each goal along the diagonal can be reached via two distinct plans. On average, the agent takes approximately 80 steps to complete an episode in both environments.

\textbf{Parking} From the environments in the highway-env package, we focus on the parking continuous control task where the ego-vehicle must park in a designated space with the correct heading. Each episode begins with the agent positioned randomly in the parking lot, adding complexity as optimal paths to the same goal can vary significantly from one episode to another. On average, an episode in this environment consists of 15 steps.

\textbf{Panda-Gym} From the collection of environments provided in the panda-gym package, we focus on the Reach continuous control environment, where the Panda arm robot must reach a specified location. The agent begins each episode at the same starting position. On average, each episode in this environment consists of approximately 20 steps.

\begin{table*}[tb]
\caption{Comparison of domains, visualizations, and their specific characteristics.}
\label{tab:domains}
\scriptsize
\centering
\renewcommand{\arraystretch}{1.2} 
\begin{tabular}{m{2cm} m{2cm} m{9cm}}
    \toprule
    \textbf{Figure} & \textbf{Domain} & \textbf{Details} \\
    \midrule
    \includegraphics[width=2cm]{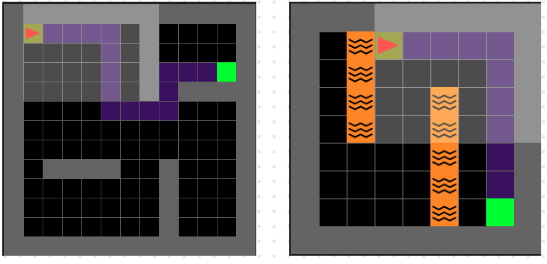} 
    & \raggedright\arraybackslash \textbf{MiniGrid} 
    & \raggedright\arraybackslash \textit{Environments}: SimpleCrossing, LavaCrossing. \newline
      \textit{Motivation}: Discrete navigation; comparison with GR baselines. \newline
      \textit{States}: Either $(x,y)$ + direction (Discrete) \emph{or} image observation + direction (Discrete). \newline
      \textit{Actions}: Turn, move forward, stay in place (Discrete). \newline
      \textit{Reward}: $1 - 0.9 \cdot (\text{step\_count} / \text{max\_steps})$ for success, else $0$ (Sparse). \\
    \midrule
    \includegraphics[width=2cm]{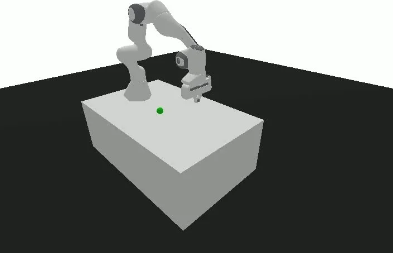} 
    & \raggedright\arraybackslash \textbf{Panda-Gym} 
    & \raggedright\arraybackslash \textit{Environments}: PandaReach. \newline
      \textit{Motivation}: Realistic 3D robotic control scenario. \newline
      \textit{States}: Robotic arm positions, velocities (Continuous). \newline
      \textit{Actions}: Joint torques for robotic arm (Continuous). \newline
      \textit{Reward}: Gradual reward based on distance to goal (Dense). \\
    \midrule
    \includegraphics[width=2cm]{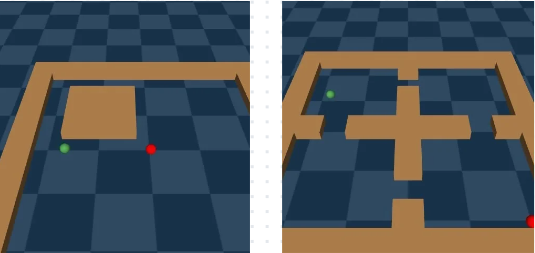} 
    & \raggedright\arraybackslash \textbf{PointMaze} 
    & \raggedright\arraybackslash \textit{Environments}: Obstacle mazes; FourRooms. \newline
      \textit{Motivation}: Continuous navigation; path complexity. \newline
      \textit{States}: Agent positions, velocities (Continuous). \newline
      \textit{Actions}: Force applied to agent (Continuous). \newline
      \textit{Reward}: $-1$ if not reaching goal; $0$ when near goal ($<0.5$) (Sparse). \\
    \midrule
    \includegraphics[width=2cm]{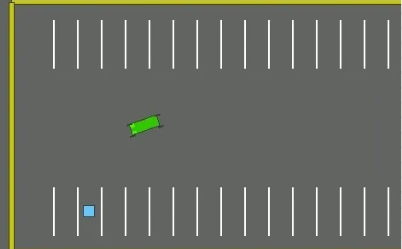}
    & \raggedright\arraybackslash \textbf{Highway-env (Parking)} 
    & \raggedright\arraybackslash \textit{Environments}: Parking (continuous control). \newline
      \textit{Motivation}: Goal-conditioned control with pose alignment (position and heading) in a cluttered parking lot. Randomized starts make optimal paths vary across episodes. \newline
      \textit{States}: KinematicsGoal features $(x,y,v_x,v_y,\cos h,\sin h)$ (Continuous). \newline
      \textit{Goal space}: Discrete (designated parking spot + heading). \newline
      \textit{Actions}: Acceleration and steering (Continuous). \newline
      \textit{Reward/Success}: Weighted $p$-norm distance to goal with collision penalty; success upon reaching the correct pose within tolerance (Sparse/Dense hybrid). \\
    \bottomrule
\end{tabular}
\end{table*}

\section{Converting Gymnasium Optimization Environment to ODGR Environment}
\label{app:convert}

Gymnasium is a standardized Python toolkit for developing and benchmarking algorithms across a wide range of environments modeled as MDPs~\cite{towers2024gymnasium}. It provides a unified API and diverse domains ranging from classic control to grid-based navigation, allowing evaluation of learning and decision-making models in simulated settings. Each environment has its own state and action representations, but they share the same API functions. These functions, however, do not provide several operations required by GR algorithms, and externally modifying environment properties is not always straightforward.

\texttt{gr-envs} addresses these limitations by providing custom Gymnasium environments and wrappers that add the functionality required by GR algorithms. It explicitly registers six different types of goal-independent environments. Each type can be instantiated with many different goals, resulting in a large number of individual environment instances.

We now elaborate on the specific requirements of GR algorithms and how \texttt{gr-envs} supports them:

\begin{description}
\item [Goal-Conditioned Adaptability] Some GR algorithms rely on training GCRL agents~\cite{liu2022goalconditioned} to optimize adaptation in dynamic environments~\citep{shamir2025gramldynamicgoalrecognition, elhadad2025generaldynamicgoalrecognition, nageris2024goal}. Not all Gymnasium environments are suitable for GCRL training. We define the property of an environment being suitable for such training as \textit{goal-conditioned adaptability}, which indicates whether a GCRL agent can be trained effectively. Table \ref{tab:envs} shows the value of this property for every supported environment in \texttt{gr-envs}. To be goal-conditioned adaptable, the environment must allow control over the contents of the goal space, either at construction or via setters, and must render goal locations in each episode from this predefined goal set or sample them from a general default goal space, enabling RL algorithms to train over trajectories that reach all these goals. Additionally, the state space must include both the desired goal and the achieved goal in the observation returned by the environment, which is the default in some environments, such as Parking, Panda, and PointMaze, but not in Minigrid. \texttt{gr-envs} provides initialization methods, environment wrappers, or custom implementations that enable control over the initial goal set and its sampling method at each environment reset. The sampling can be random, drawn from different goal regions, or uniform over a discrete set depending on the domain. \texttt{gr-envs} also ensures that the desired goal and the achieved goal are returned as part of the observation when performing a step in the environment.

\item [Goal Control] Assuming the environment is goal-conditioned adaptable, the goal is randomized from the specified goal set at each environment reset. However, unlike GCRL algorithms, GR algorithms also require the ability to explicitly control the goal location for other purposes, such as extracting goal-directed policies from a goal-conditioned policy~\cite{amado2022goal, nageris2024goal, elhadad2025generaldynamicgoalrecognition} or generating goal-directed sequences~\cite{shamir2025gramldynamicgoalrecognition}. Goal control is also necessary for creating benchmark GR problems and providing inputs to the inference phase. Some environments in \texttt{gr-libs} do not natively provide this functionality. For instance, PointMaze allows specifying the goal location in its constructor, which enables generating different environments with different goal sets for extracting goal-directed policies. In contrast, Parking, Panda-Reach, and Minigrid do not provide this capability through API functions or constructors. \texttt{gr-envs} adapts these environments for GR by generating custom variants with explicit goal specification, such as MyReach for Panda and DynamicFourRoomsEnv and DynamicGoalCrossingEnv for Minigrid, or by providing wrappers that enable goal manipulation, such as ParkingWrapper, PandaGymWrapper, and PointMazeWrapper.
\end{description}

\begin{table}[ht]
\centering
\scriptsize
\caption{Comparison of supported environments in the GR benchmark. GC-adaptability is not yet supported in minigrid environments.}
\label{tab:envs}
\begin{tabularx}{\textwidth}{lXXX}
\toprule
\textbf{Environment} & \textbf{State Compatibility} & \textbf{Action Compatibility} & \textbf{GC Adaptable} \\
\midrule
Minigrid-SimpleCrossing & Discrete & Discrete & \xmark \\
Minigrid-LavaCrossing & Discrete & Discrete & \xmark \\
PointMaze-Obstacle & Continuous & Continuous & \cmark \\
PointMaze-FourRooms & Continuous & Continuous & \cmark \\
Parking (highway-env) & Continuous & Continuous & \cmark \\
Panda-Gym Reach & Continuous & Continuous & \cmark \\
\bottomrule
\end{tabularx}
\end{table}

\section{Benchmark Evaluation}
\label{app:eval}

From the parallel execution of ODGR problems in the script \texttt{all$\_$experiments.py}, detailed and compiled versions of the results are produced, from which comparison tables can be made. After running the script on the supported environments, we get a baseline evaluation for researchers to compare to, as seen in Table \ref{tab:odgr_all}.

\begin{table}[htbp]
\centering
\renewcommand{\arraystretch}{1.3}
\setlength{\tabcolsep}{3pt}
\caption{Accuracy of 5 GR algorithms across four environments (PandaReach, Parking, PointMaze-FourRooms, PointMaze-Obstacles) and multiple observability levels, using consecutive (top) and non-consecutive (bottom) observation sequences. Algorithms: A = GCGraml, B = GCAura, C = GCDraco, D = Draco, E = BGGraml.}

\begin{adjustbox}{width=1\textwidth, center}
\begin{tabular}{l*{20}{c}}
\toprule
\multirow{3}{*}{\textbf{Environment}} 
  & \multicolumn{20}{c}{\textbf{Consecutive Observations}} \\
\cmidrule(lr){2-21}
  & \multicolumn{5}{c}{0.3} & \multicolumn{5}{c}{0.5} & \multicolumn{5}{c}{0.7} & \multicolumn{5}{c}{1.0} \\
\cmidrule(lr){2-6} \cmidrule(lr){7-11} \cmidrule(lr){12-16} \cmidrule(lr){17-21}
  & A & B & C & D & E & A & B & C & D & E & A & B & C & D & E & A & B & C & D & E \\
\midrule
PandaReach & 0.8189 & 0.8667 & 0.8622 & 0.9000 & 0.7900 & 0.9822 & 0.8444 & 0.8444 & 0.8600 & 0.9500 & 0.9644 & 0.9556 & 0.9556 & 0.9700 & 0.9400 & 0.9022 & 0.9778 & 0.9778 & 0.9800 & 0.8800 \\
Parking    & 0.5410 & 0.5292 & 0.5376 & 1.0000 & 0.9300 & 0.6228 & 0.5292 & 0.5376 & 0.9500 & 0.8775 & 0.8381 & 0.5292 & 0.5376 & 1.0000 & 0.7842 & 0.8454 & 0.5292 & 0.5376 & 0.9500 & 0.9175 \\
PointMaze-FourRooms & 0.679 & 0.219 & 0.219 & 0.559 & 0.751 & 0.736 & 0.219 & 0.219 & 0.792 & 0.798 & 0.852 & 0.219 & 0.219 & 0.775 & 0.907 & 0.970 & 0.219 & 0.219 & 0.865 & 0.959 \\
PointMaze-Obstacles & 0.360 & 0.254 & 0.257 & 0.573 & 0.404 & 0.515 & 0.254 & 0.257 & 0.808 & 0.416 & 0.733 & 0.254 & 0.257 & 0.915 & 0.701 & 0.597 & 0.254 & 0.257 & 0.872 & 0.795 \\
\bottomrule
\end{tabular}
\end{adjustbox}

\vspace{1em} 

\begin{adjustbox}{width=1\textwidth, center}
\begin{tabular}{l*{20}{c}}
\toprule
\multirow{3}{*}{\textbf{Environment}} 
  & \multicolumn{20}{c}{\textbf{Non-Consecutive Observations}} \\
\cmidrule(lr){2-21}
  & \multicolumn{5}{c}{0.3} & \multicolumn{5}{c}{0.5} & \multicolumn{5}{c}{0.7} & \multicolumn{5}{c}{1.0} \\
\cmidrule(lr){2-6} \cmidrule(lr){7-11} \cmidrule(lr){12-16} \cmidrule(lr){17-21}
  & A & B & C & D & E & A & B & C & D & E & A & B & C & D & E & A & B & C & D & E \\
\midrule
PandaReach & 0.8689 & 0.8667 & 0.8667 & 0.8800 & 0.8400 & 0.9756 & 0.9778 & 0.9778 & 1.0000 & 0.9500 & 0.9178 & 0.9778 & 0.9778 & 1.0000 & 0.8900 & 0.9022 & 0.9778 & 0.9778 & 1.0000 & 0.8800 \\
Parking    & 0.8820 & 0.5292 & 0.5376 & 0.9125 & 0.8858 & 0.9003 & 0.5292 & 0.5376 & 0.9800 & 0.8883 & 0.8614 & 0.5292 & 0.5376 & 0.9600 & 0.9175 & 0.8709 & 0.5292 & 0.5376 & 0.9500 & 0.9175 \\
PointMaze-FourRooms & 0.983 & 0.219 & 0.219 & 0.820 & 0.973 & 0.986 & 0.219 & 0.219 & 0.844 & 0.966 & 0.984 & 0.219 & 0.219 & 0.860 & 0.956 & 0.992 & 0.219 & 0.219 & 0.858 & 0.941 \\
PointMaze-Obstacles & 0.695 & 0.254 & 0.257 & 0.851 & 0.750 & 0.722 & 0.254 & 0.257 & 0.889 & 0.771 & 0.729 & 0.254 & 0.257 & 0.873 & 0.736 & 0.647 & 0.254 & 0.257 & 0.883 & 0.779 \\
\bottomrule
\end{tabular}
\end{adjustbox}

\label{tab:odgr_all}
\end{table}

\end{document}